\PassOptionsToPackage{unicode}{hyperref}
\PassOptionsToPackage{hyphens}{url}
\documentclass[
  11pt,
]{article}
\usepackage{xcolor}
\usepackage[margin=1in]{geometry}
\usepackage{amsmath,amssymb}
\usepackage{iftex}
\ifPDFTeX
  \usepackage[T1]{fontenc}
  \usepackage[utf8]{inputenc}
  \usepackage{textcomp} 
\else 
  \usepackage{unicode-math} 
  \defaultfontfeatures{Scale=MatchLowercase}
  \defaultfontfeatures[\rmfamily]{Ligatures=TeX,Scale=1}
\fi
\usepackage{lmodern}
\ifPDFTeX\else
\fi
\IfFileExists{upquote.sty}{\usepackage{upquote}}{}
\IfFileExists{microtype.sty}{
  \usepackage[]{microtype}
  \UseMicrotypeSet[protrusion]{basicmath} 
}{}
\makeatletter
\@ifundefined{KOMAClassName}{
  \IfFileExists{parskip.sty}{%
    \usepackage{parskip}
  }{
    \setlength{\parindent}{0pt}
    \setlength{\parskip}{6pt plus 2pt minus 1pt}}
}{
  \KOMAoptions{parskip=half}}
\makeatother
\usepackage{longtable,booktabs,array}
\usepackage{calc} 
\usepackage{etoolbox}
\makeatletter
\patchcmd\longtable{\par}{\if@noskipsec\mbox{}\fi\par}{}{}
\makeatother
\IfFileExists{footnotehyper.sty}{\usepackage{footnotehyper}}{\usepackage{footnote}}
\makesavenoteenv{longtable}
\usepackage{graphicx}
\makeatletter
\newsavebox\pandoc@box
\newcommand*\pandocbounded[1]{
  \sbox\pandoc@box{#1}%
  \Gscale@div\@tempa{\textheight}{\dimexpr\ht\pandoc@box+\dp\pandoc@box\relax}%
  \Gscale@div\@tempb{\linewidth}{\wd\pandoc@box}%
  \ifdim\@tempb\p@<\@tempa\p@\let\@tempa\@tempb\fi
  \ifdim\@tempa\p@<\p@\scalebox{\@tempa}{\usebox\pandoc@box}%
  \else\usebox{\pandoc@box}%
  \fi%
}
\def\fps@figure{htbp}
\makeatother
\providecommand{\tightlist}{%
  \setlength{\itemsep}{0pt}\setlength{\parskip}{0pt}}
\usepackage{bookmark}
\IfFileExists{xurl.sty}{\usepackage{xurl}}{} 
\makeatletter
\@ifundefined{xmpquote}{}{}
\makeatother
\hypersetup{
  pdftitle={Detectable Only Where It Is Confounded: What Verified Duplication Counts Say About Membership Evidence in Language Models},
  pdfauthor={Arman Nik Khah, The University of Texas at Dallas},
  hidelinks,
  pdfcreator={LaTeX via pandoc}}

\title{Detectable Only Where It Is Confounded: What Verified Duplication
Counts Say About Membership Evidence in Language Models}
\author{Arman Nik Khah, The University of Texas at Dallas}
\date{September 2026}

\begin{document}
\maketitle

\section{Abstract}\label{abstract}

When a language model finds a sentence unusually cheap to predict, it is
tempting to conclude that the sentence was in its training data. Almost
every published test of that inference has had to guess which sentences
were in the training data, the members, and which were not. This paper
removes the guessing. Two model families, OLMo-2 and Pythia, publish
their pretraining corpora, and a public index over those corpora returns
the exact number of times any sentence appeared in each. Those counts
make three questions answerable directly. How many copies does it take
before a model's loss carries a trace of exposure? Can that trace be
separated from the ordinary fact that some sentences are easier than
others? And how much of the reported success of membership tests comes
from the way their control sentences were built?

The answers form a pincer, closing from two sides. At the duplication
levels ordinary text actually has, five models from 1B to 13B parameters
carry at most a faint trace of their own exposure. We measure that trace
with a design that reads the same sentence through two models, which
cancels fluency and quality by construction, and it comes to a rank
correlation near $-$0.08, where $-$1 would be a perfect relation and 0 none.
Where the trace does become strong, above roughly a thousand copies, the
two corpora agree on which sentences those are, because they are the
famous ones, so exposure can no longer be told apart from fame. Two
further measurements show how apparent membership signal gets
manufactured. A common way to build a non-member is to change one word
of a member. The model does prefer the original, but the gap is the
same, within noise, whether the original appeared once or a hundred
times, so what the model is rewarding is the author's word choice, not
memory. Above a thousand copies the gap grows with model size on the
twelve sentences we can test there, at the same boundary where the
pincer closes. And swapping the controls for sentences that differ from
the members in register moves a detector from 0.83 to 0.94 AUC, on a
scale where 0.5 is a coin flip and 1.0 is perfect separation. We release
the sentence banks, counts, and code.

\section{1. The question, and why it has been hard to
answer}\label{the-question-and-why-it-has-been-hard-to-answer}

Suppose someone hands you a language model and a sentence, and asks
whether the model was trained on that sentence. There is an obvious
place to look. Language models are trained to make their training text
cheap to predict, so a sentence the model has seen should cost it less,
in loss, than a comparable sentence it has not. That intuition is what
membership inference rests on, a family of tests that read a model's
loss, or a statistic derived from it, and answer ``member'' or
``non-member'' {[}Shi et al., 2024; Mattern et al., 2023{]}.

The intuition is sound. The tests built on it have not worked well. A
careful evaluation across model sizes and domains found that most
membership tests on large language models barely beat a coin flip
{[}Duan et al., 2024{]}, and a systematic review argued that the tests
which do appear to work owe much of their success to how their
evaluation sets were built {[}Meeus et al., 2024{]}. When members and
non-members are assembled after the fact, by guessing what a model
probably saw, the two sets differ in date, source, style, and length,
and a detector can score well by noticing any of those differences
without noticing membership at all. Baselines that never touch the model
at all, and classify on date or style alone, can match published attacks
on exactly this basis {[}Das et al., 2024{]}.

Two things have been missing from this conversation. The first is a
number. Membership is not a binary property of web-scale training. A
sentence from a novel may appear in a corpus once, or fifty times, or
ten thousand times, because the web contains many copies of many things.
Memorization scales with that count {[}Carlini et al., 2022{]}, and
detectors of memorized text are near chance on sequences that appear
only once {[}Kandpal et al., 2022{]}. But membership benchmarks score
tests against a yes-or-no label, which cannot say at what count the loss
signal appears. The second missing thing is a way to vary exposure while
holding everything else fixed. Every member-versus-non-member comparison
compares different sentences, and different sentences differ in more
than membership.

This paper supplies both. Two open model families publish their complete
pretraining data. OLMo-2 was trained on a corpus called OLMo-mix-1124
{[}OLMo Team, 2025{]}, and Pythia (the standard models, not the
deduplicated ones) was trained on the Pile {[}Biderman et al., 2023; Gao
et al., 2020{]}. A public index, infini-gram {[}Liu et al., 2024{]},
returns the exact number of times any token sequence occurs in either
corpus, in milliseconds. So for any sentence we can state, not guess,
how many times each family saw it. That turns membership from a label
into a dose. And because the two corpora were assembled independently,
the same sentence can be common in one and rare in the other. So we can
vary exposure while holding the sentence, and therefore its fluency,
quality, and length, exactly fixed. We read the sentence through a model
from each family, and ask whether the family that saw it more finds it
cheaper.

The findings, in the order the paper presents them:

\begin{enumerate}
\def\labelenumi{\arabic{enumi}.}
\tightlist
\item
  \textbf{Loss-based membership is flat within noise from 1 to 1,000
  copies and turns on above that.} Against non-members built from the
  same sentences, a 1B model scores 0.60 AUC on ordinary sentences and
  0.83 on famous lines with a median of about 1,200 copies (0.5 is
  chance, 1.0 is perfect separation).
\item
  \textbf{Where the two corpora disagree about exposure, the trace of
  exposure is faint.} In the same-sentence design, comparing each
  sentence only with others from the same book, the model that saw a
  sentence more finds it cheaper with rank correlation $-$0.08 at 7B and
  $-$0.07 at 13B, on 747 sentences. That is under one percent of the
  variance, and the effect does not grow with size.
\item
  \textbf{A one-word edit is not an equal-quality non-member.} In over
  ninety percent of pairs the original costs the model less than its
  edit, by about 0.4 nats per token, which makes the original's tokens
  about 1.5 times as probable on average. The gap is the same, within
  noise, for an original that appeared once as for one that appeared a
  hundred times, so it measures the author's word choice, not memory.
  Above a thousand copies the gap does grow with model size, so only
  below that does the gap measure word choice alone.
\item
  \textbf{The control set does the rest.} On identical member sentences,
  swapping same-source edits for composed-prose controls moves a
  detector from 0.83 to 0.94.
\end{enumerate}

Findings 1 and 2 are the pincer. Exposure leaves a measurable trace only
at counts where both corpora agree on which sentences are exposed, and
the corpora disagree only at counts where the trace is faint. Section 8
states this precisely and says what would break it, and Figure 6 shows
how far count and the choice of controls move one detector.

\section{2. Counting exposure}\label{counting-exposure}

\subsection{2.1 The instrument}\label{the-instrument}

Infini-gram {[}Liu et al., 2024{]} is a suffix-array index over a
tokenized corpus. Given a token sequence, it returns the number of
positions in the corpus at which that exact sequence occurs. We use two
of its public indexes. One covers OLMo-mix-1124, the pretraining corpus
of the OLMo-2 models, and the other covers the training split of the
Pile, the pretraining corpus of the Pythia models. A query costs about
thirty milliseconds and requires no key. Throughout, ``copies'' means
this exact-match count.

Three properties of the count matter for everything that follows.

\emph{It is exact, so it is a floor.} A sentence reprinted with
different punctuation, a curly quote, or a line break in a different
place is a different token sequence and is not counted. The count
therefore understates exposure, never overstates it. The OLMo side is a
floor for a second reason. All three OLMo-2 models we use finish
pretraining with a short annealing stage on a separate mix that the
index does not cover {[}OLMo Team, 2025{]}, so any exposure there is
uncounted.

\emph{A zero needs a presence check.} An index that is down,
rate-limited, or given a malformed query also returns nothing, and
nothing looks like zero. Every batch of queries in this paper therefore
begins with a sentence known to be present (``It is a truth universally
acknowledged'') and refuses to write a bank if that sentence returns
zero. This check is distinct from the positive control of Section 4,
which is a bank where the effect is known to exist.

\emph{The unit is a sentence, not a paragraph.} Figure 1 shows how the
count behaves as a span is extended word by word through a passage. For
most spans it is stable out to forty words, which means the copies in
the corpus are whole copies of the book. For a minority it collapses by
an order of magnitude somewhere past twenty-five words, where the copies
begin to differ in formatting, and for one it falls to zero by ten
words. A 140-word paragraph typically returns zero. We therefore work
with sentences of 10 to 16 words, long enough to be specific to one book
and short enough to survive exact matching.

\begin{figure}
\centering
\pandocbounded{\includegraphics[keepaspectratio,alt={Figure 1. Exact-match copies in OLMo-mix along twelve mid-book spans from six novels, as each span is extended from 6 to 40 words. Most spans hold their count; a minority collapse past 25 words, and one falls to zero by 10. Zero is drawn at the dotted line.}]{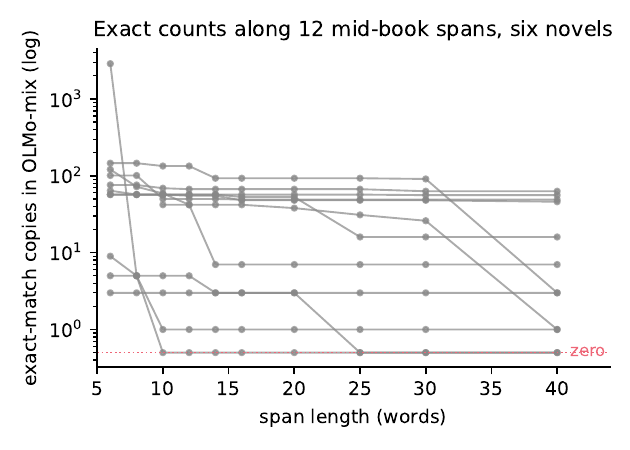}}
\caption{Exact-match copies in OLMo-mix along twelve mid-book
spans from six novels, as each span is extended from 6 to 40 words. Most
spans hold their count; a minority collapse past 25 words, and one falls
to zero by 10. Zero is drawn at the dotted line.}
\end{figure}

\subsection{2.2 The two corpora barely
agree}\label{the-two-corpora-barely-agree}

If the two corpora exposed every sentence in the same proportion,
reading a sentence through one model from each family would tell us
nothing. They do not. Across the 747 member sentences that carry both
counts, the rank (Spearman) correlation between the OLMo-mix count and
the Pile count is only 0.15. The ratio of the two counts varies by a
factor of about twenty-seven between a sentence at the tenth percentile
and one at the ninetieth. In log10 that is a span of 1.44, from $-$0.36 to
+1.08. Most of that spread is between books. A mid-book sentence from
\emph{Moby Dick} appeared about a hundred times in OLMo-mix and about
ten times in the Pile in our first bank, while a sentence from the
\emph{Sherlock Holmes} stories appeared five times and fourteen. Section
5 uses that spread as a lever to vary exposure without changing the
sentence. The between-book part of it, meaning how much the six novels
differ from each other as wholes, is the part to distrust, and Section
5.1 is where that distrust is paid off, because two model families can
differ in how easy they find a whole book for reasons that have nothing
to do with copies.

\section{3. The sentence bank}\label{the-sentence-bank}

\subsection{3.1 Where the sentences come
from}\label{where-the-sentences-come-from}

The mid-book sentences come from six public-domain novels obtained from
Project Gutenberg: \emph{Pride and Prejudice}, \emph{Moby Dick},
\emph{Frankenstein}, \emph{The Adventures of Sherlock Holmes}, \emph{The
Great Gatsby}, and \emph{Dracula}. The famous bank is drawn more widely,
because famous text is not confined to novels. Three of its twelve lines
open books on that list, and the rest come from Dickens, Tolstoy,
Carroll and Darwin, from \emph{Hamlet}, from the Gettysburg Address and
the Declaration of Independence, from the King James psalter, and from
Orwell. That spread is a feature rather than an inconsistency, because
Section 8 turns on famous text being famous everywhere, and a bank
crossing four centuries and five genres tests that harder than six
novels would. These books are certainly in both corpora, in many copies.
That is what makes them useful, because they let exposure vary across a
wide range without any doubt about the source.

Sentences are drawn from the middle of each book, skipping the first and
last tenth. Skipping the ends is the single most consequential choice in
the design. The famous opening lines of these novels are among the most
quoted text in existence, and they are also, for that reason, among the
lowest-loss text a model ever sees. A bank built from opening lines
gives a detector a nearly saturated baseline before any second signal is
measured. Our own first attempt did exactly that, reached 0.90 AUC on
loss alone, and taught us that a null against such a baseline is
uninformative. Mid-book sentences appear in a corpus tens of times
rather than thousands, and that is the regime a membership test would
actually face.

Candidate sentences are 10 to 16 words, contain only letters and
ordinary punctuation, and are sampled at random. Each is queried against
both indexes.

\subsection{3.2 Non-members from the same
source}\label{non-members-from-the-same-source}

A non-member should be a sentence the model did not see that is
otherwise like the members. The usual practice, drawing non-members from
a different source or a later date, is precisely what the
distribution-shift critique targets. We build most non-members from a
member instead, and Section 3.3 describes the rest. We swap one content
word for a one-word near-synonym of the same register, meaning the same
level and style of language, literary rather than plain (``house'' to
``cottage'', ``said'' to ``remarked''). The twelve famous-line edits
were written one at a time rather than drawn from a table, and two of
them change the same word twice. We then query the result against the
index and keep it only if it returns zero copies. Such a non-member is
therefore the same author, the same book, the same length, and the same
sentence but for one word, and we verify its absence from the corpus
rather than assume it. That check is exact-string, the same floor
Section 2.1 applies to the members, so it rules out the sentence
appearing verbatim and not every paraphrase of it. That verification
earns its keep. In our famous-line bank, two of twelve edits turned out
to be real variants that people had actually written, and we relabelled
them as members and left them out of the membership tests.

Section 6 will show that even this control is not as clean as it looks.
We build it carefully so that its remaining flaw can then be measured.

\subsection{3.3 Three banks}\label{three-banks}

Three banks are used. Pass 3 and pass 4 are the third and fourth
candidate samples drawn in this project, and the numbering is kept so
that the released files match the text.

{\def\LTcaptype{none} 
\begin{longtable}[]{@{}
  >{\raggedright\arraybackslash}p{(\linewidth - 8\tabcolsep) * \real{0.2000}}
  >{\raggedright\arraybackslash}p{(\linewidth - 8\tabcolsep) * \real{0.2000}}
  >{\raggedright\arraybackslash}p{(\linewidth - 8\tabcolsep) * \real{0.2000}}
  >{\raggedright\arraybackslash}p{(\linewidth - 8\tabcolsep) * \real{0.2000}}
  >{\raggedright\arraybackslash}p{(\linewidth - 8\tabcolsep) * \real{0.2000}}@{}}
\toprule\noalign{}
\begin{minipage}[b]{\linewidth}\raggedright
bank
\end{minipage} & \begin{minipage}[b]{\linewidth}\raggedright
source
\end{minipage} & \begin{minipage}[b]{\linewidth}\raggedright
members
\end{minipage} & \begin{minipage}[b]{\linewidth}\raggedright
non-members
\end{minipage} & \begin{minipage}[b]{\linewidth}\raggedright
use
\end{minipage} \\
\midrule\noalign{}
\endhead
\bottomrule\noalign{}
\endlastfoot
famous & opening lines and well-known quotations & 12 verbatim & 10
same-source edits, 12 composed prose & positive control; control-set
comparison \\
pass 3 & mid-book, six novels & 162 & 102 same-source edits, 13
zero-count sentences & baseline by count band; edit confound \\
pass 4 & mid-book, six novels, fresh sample & 585 & 49 zero-count
sentences & differential design, combined with pass 3 \\
\end{longtable}
}

The famous bank has a median of 1,192 OLMo-mix copies among its members,
and the mid-book banks have medians in the tens. For the differential
design, members from pass 3 and pass 4 are pooled, giving 747 sentences
with both counts, and every model pair is scored on the same 747. The
mid-book banks also hold a second kind of non-member, a sampled sentence
that the OLMo-mix index returns zero times. There are 13 of these among
the 115 non-members of pass 3, and all 49 non-members of pass 4 are of
this kind. Their absence is also exact-string only, and their books are
in the corpus. Of the 102 edits, 92 change one word of a member and 10
change one word of a sampled sentence that is not a labelled member;
Section 6 pairs the 92 with their members. Every pass-3 AUC labelled as
against same-source edits uses all 115. At the median, the 13 cost
OLMo-2 1B less than the members do, and without them every pass-3 score
rises by about 0.03 (Appendix B).

\section{4. Where loss-based membership turns
on}\label{where-loss-based-membership-turns-on}

The simplest membership test is the loss itself. Call a sentence a
member if the model's mean per-token loss on it is low. We score every
sentence in a bank under OLMo-2 1B and report the area under the ROC
curve, with members as positives, so that 0.5 is chance and 1.0 is
perfect separation.

On the full pass-3 bank the loss test scores 0.60 AUC (95 percent
bootstrap interval 0.56 to 0.64). Figure 2 breaks the members into bands
by copy count. Members with 1 to 9 copies score 0.66, members with 10 to
99 copies score 0.59, and members with 100 to 999 copies score 0.57. The
intervals overlap heavily, and the ordering runs the wrong way for a
memorization story, because the lowest band scores highest. Loss-based
membership is therefore flat within noise across three orders of
magnitude of duplication.

Against the same kind of control, the famous bank scores 0.83 (interval
0.65 to 0.98, twelve members against the ten edits that Section 3.2
kept). The interval is wide because the bank is small, so treat it as a
positive control rather than a precise estimate. It still establishes
what the flat region does not, which is that the instrument can see
exposure, or whatever travels with it, when there is enough of it.
Enough lies somewhere above the hundreds and around the low thousands.

\begin{figure}
\centering
\pandocbounded{\includegraphics[keepaspectratio,alt={Figure 2. AUC of the loss test against same-source edits, by median copy count of the members. Blue: mid-book sentences in three count bands. Red: famous lines. Bars are 95 percent bootstrap intervals; the dashed line is chance.}]{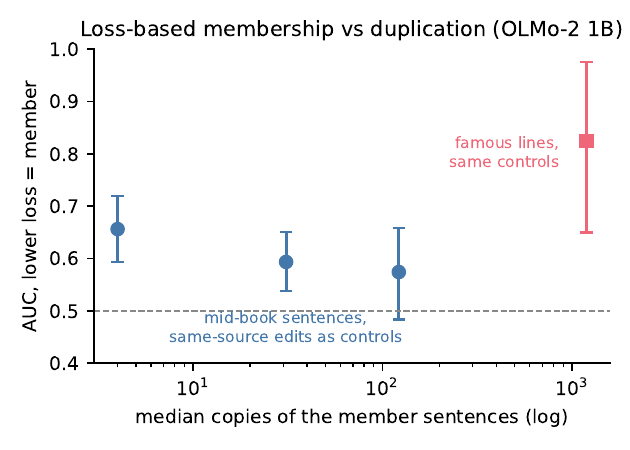}}
\caption{AUC of the loss test against same-source edits, by
median copy count of the members. Blue: mid-book sentences in three
count bands. Red: famous lines. Bars are 95 percent bootstrap intervals;
the dashed line is chance.}
\end{figure}

\section{5. The differential design}\label{the-differential-design}

\subsection{5.1 What it cancels, and what it does
not}\label{what-it-cancels-and-what-it-does-not}

Every number in Section 4 compares different sentences. A member and its
edit differ by one word, but they still differ, and Section 6 will show
that the one word is doing work. The differential design removes the
comparison between sentences entirely.

Take one sentence. Read it through an OLMo-2 model and through a Pythia
model of similar size, and record the difference in their mean per-token
loss. Now look up how many times OLMo-mix contained the sentence and how
many times the Pile did, and record the difference in the logs of those
counts. If exposure leaves a trace, the family that saw the sentence
more should find it cheaper, so across sentences the loss difference
should fall as the count difference rises. Fluency, quality, length, and
topic are properties of the sentence, and the sentence is the same on
both sides of the subtraction, so they cancel as far as both families
weigh them alike. The two families also tokenize text differently, which
the subtraction does not remove.

One thing does not cancel. The count ratio varies mostly between books
(Section 2.2), and two model families may find a given author's prose
easier or harder for reasons that have nothing to do with copies, such
as the composition of their corpora, the era of the text, or the
register. Suppose OLMo-2 finds Melville generally easier than Pythia
does, and Melville also happens to be the author with the highest
OLMo-to-Pile ratio. A pooled correlation would then report exposure
where there is only taste. So we run the test within books. That means
each sentence is compared only with other sentences from the same novel.
Among the sentences of \emph{Moby Dick}, are the ones OLMo-mix saw more
the ones OLMo-2 finds cheaper? The same question is asked inside each of
the other five books, and the answers are pooled. The between-book
question is the other one. It treats each novel as a single point and
asks whether the six points line up, and that is where taste can hide.
We centre each sentence's count difference and loss difference on its
book's mean and compute the rank correlation on the centred values. To
assess significance we shuffle the loss differences within each book
five thousand times and ask how often a shuffle produces a correlation
at least as large as the real one. We report the between-book
correlation separately, on six book medians, and do not treat it as
evidence about exposure.

\subsection{5.2 Results}\label{results}

Figure 3 shows the design for two size pairs, OLMo-2 13B against Pythia
12B and OLMo-2 7B against Pythia 6.9B. The left panels are the
within-book test and the right panels are the six book medians. The left
clouds look like noise, which is what a correlation this small looks
like. An effect worth under one percent of the variance is not visible
to the eye, which is the reason the test is a permutation test and not a
look.

Within books, the correlation has the predicted sign in both pairs and
is small in both. For the 7B pair it is $-$0.084 (permutation p = 0.016, n
= 747). For the 13B pair it is $-$0.065 (permutation p = 0.067, n = 747).
Squaring a rank correlation of $-$0.08 gives about 0.007, so exposure
accounts for under one percent of the variance in the loss difference.
That $-$0.08 is the paper's central number. At the counts where two public
corpora disagree about exposure, exposure leaves a trace that is real at
7B, marginal at 13B, and faint at both.

Between books the correlation is large, and it is the number to be
suspicious of. It is $-$0.94 for the 7B pair and $-$0.49 for the 13B pair,
on six points. A pooled analysis would have reported that effect, and
Figure 3 shows what it is made of. In the 7B pair, \emph{Moby Dick} and
\emph{The Great Gatsby} sit at the high-ratio, low-difference end, and
\emph{Sherlock Holmes} and \emph{Dracula} sit at the other. With six
books there is no way to decide whether that pattern is book-level
memorization or a difference in taste between the two families.

The one test available leans toward taste. If the between-book pattern
were memorization, it should strengthen with model size, since
memorization does {[}Carlini et al., 2022{]}. It does not. On the 585
sentences of the 747 that come from pass 4, scored under both pairs, the
between-book correlation is $-$0.89 at 7B and $-$0.43 at 13B, so the point
estimate moves the wrong way for memorization. The 13B-minus-7B
difference is +0.46, and a bootstrap over sentences within books puts
its 95 percent interval from $-$0.63 to +0.74. The evidence is compatible
with no size trend and does not support a growing one.

The same-sentence design therefore finds a real but faint trace of
exposure within books, alongside a large between-book pattern that the
size test suggests is taste rather than memory.

\begin{figure}
\centering
\pandocbounded{\includegraphics[keepaspectratio,alt={Figure 3. The differential design. Left: for each sentence, the OLMo-minus-Pythia loss difference against the OLMo-minus-Pile log count difference (natural log of count plus one), both centred on the sentence's book; the line is a least-squares fit for the eye; r and p are the within-book Spearman correlation and its within-book permutation p.~Right: the same quantities as medians per book. Top row: 13B vs 12B. Bottom row: 7B vs 6.9B.}]{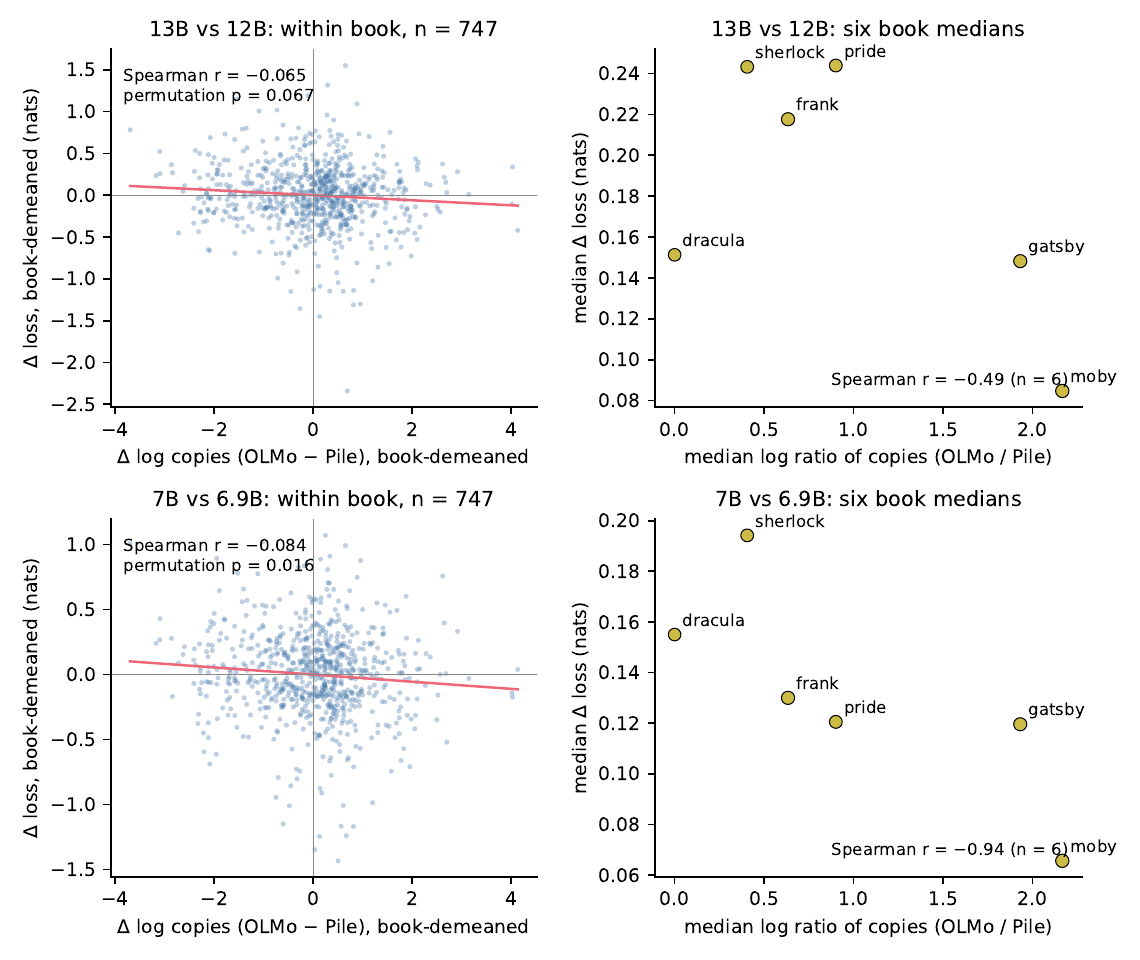}}
\caption{The differential design. Left: for each sentence, the
OLMo-minus-Pythia loss difference against the OLMo-minus-Pile log count
difference (natural log of count plus one), both centred on the
sentence's book; the line is a least-squares fit for the eye; r and p
are the within-book Spearman correlation and its within-book permutation
p.~Right: the same quantities as medians per book. Top row: 13B vs 12B.
Bottom row: 7B vs 6.9B.}
\end{figure}

Figure 4 places these results on a size ladder. Its left panel asks the
raw question, without the differential design. Does a model's own loss
track its own corpus count, across members? Pooled over all six books,
the answer splits by family. The three OLMo-2 models come out slightly
positive, at 0.047, 0.056, and 0.015, meaning more copies went with
marginally higher loss, and none is significant. The two Pythia models
come out negative, at $-$0.058 and $-$0.079, and the larger of the two,
Pythia 12B, reaches p = 0.03. That split is the confound of Section 5.1
caught in the act, and centring on book removes it. Within books, the
four models that can be centred all land between $-$0.016 and $-$0.056, the
predicted sign, and none is significant (OLMo-2 7B $-$0.040, OLMo-2 13B
$-$0.056, Pythia 6.9B $-$0.016, Pythia 12B $-$0.040; the 1B scores carry no
book label and stay pooled). The hollow bars in the figure show those
centred values. Read within book, a model's own loss carries the same
faint trace of its own exposure that the differential design finds, and
the family split was book taste. The right panel repeats the within-book
differential estimates with bootstrap intervals, and the trace does not
grow from 7B to 13B. One caveat attaches to that comparison. The
13B-class models were run in 8-bit integer precision to fit a 24 GB GPU,
while the 7B-class models ran in half precision. Quantization makes each
loss measurement noisier, and noise pulls correlations toward zero, so a
flat-to-shrinking trend at 13B could partly be a measurement artifact
rather than a fact about the models. An fp16 run of the 13B pair on a
larger card would settle it.

\begin{figure}
\centering
\pandocbounded{\includegraphics[keepaspectratio,alt={Figure 4. Left: Spearman correlation between a model's own-corpus copy count and its loss, across member sentences, for five models. Filled bars pool all six books; hollow bars centre each sentence on its book, which removes the taste confound of Section 5.1 (the 1B scores carry no book label). Permutation p at each filled bar's tip. Right: the within-book differential correlation for the two size pairs, with bootstrap intervals and permutation p.}]{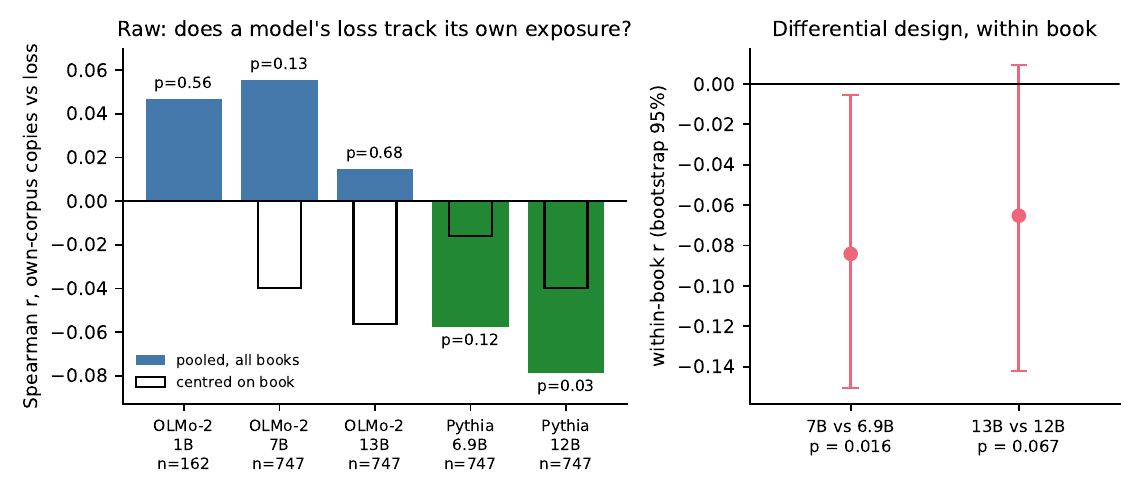}}
\caption{Left: Spearman correlation between a model's
own-corpus copy count and its loss, across member sentences, for five
models. Filled bars pool all six books; hollow bars centre each sentence
on its book, which removes the taste confound of Section 5.1 (the 1B
scores carry no book label). Permutation p at each filled bar's tip.
Right: the within-book differential correlation for the two size pairs,
with bootstrap intervals and permutation p.}
\end{figure}

\section{6. Edit controls manufacture
membership}\label{edit-controls-manufacture-membership}

Return to the same-source edit, the control that Section 3.2 built
carefully. In pass 3, 92 of the 162 members have one, a partner that
differs from the member by one content word and is verified to occur
zero times in the corpus. If loss saw membership, the partner should
cost more than the original, and the gap should grow with the original's
copy count, because the original is more memorized and the partner is
not memorized at all.

Figure 5 shows the gap. The original beats its edit by a median of about
0.4 nats per token, in 92 to 100 percent of pairs, at every model size.
A median gap of 0.4 nats means that, averaged over its tokens, the
original is about 1.5 times as probable as the edit; individual tokens
vary widely around that. If the gap were memory it would grow with
copies. At 1B and 7B it does not, and at 13B it rises from 0.332 at 1 to
9 copies to 0.568 at 100 to 999, with intervals that overlap (Figure 5).
What it measures is fit. The original author's word suits the sentence
better than our near-synonym does. ``Quite a week'' costs less than
``rather a week'' because Austen chose the first, and the model, having
read a great deal of English, agrees with her.

The flatness has an edge, and it falls where the rest of this paper says
it should. Run the same paired measurement on the famous bank, whose
members carry a median of 1,192 copies, and the gap stops being flat. At
13B it reaches 0.932 nats, against 0.568 in the 100 to 999 band. The
same twelve pairs, the two real variants included, give 0.550 at 1B and
0.787 at 7B, so the gap grows with size across all three models. The 7B
scores are half precision, which matters here, because it means the
growth cannot be an artifact of the 8-bit arithmetic the 13B models
needed. Fit accounts for that growth less well than memory does, because
no mid-book band grows as much from 1B to 13B, and memorization grows
with model size {[}Carlini et al., 2022{]}. Section 8 returns to this
contrast. So the edit control measures word choice across the range
ordinary text occupies, and begins measuring something else above about
a thousand copies, which is the boundary Section 4 found and Section 8
turns into the argument.

This has a direct consequence for evaluation. The unpaired AUC on the
mid-book bank was 0.60 (Section 4). A paired test, asking for each pair
whether the original costs less than its edit, would score above 0.9 on
the same data. That number would look like a strong membership detector.
It would be a fluency detector, and it would score equally well on
sentences the model had seen once. Any evaluation that builds its
non-members by minimally perturbing members inherits the same problem,
because the perturbation is detectable in its own right, and detecting
it is not detecting membership. Neighbourhood attacks {[}Mattern et al.,
2023{]} perturb members and non-members alike, so the fit gap enters
both sides of their comparison.

\begin{figure}
\centering
\pandocbounded{\includegraphics[keepaspectratio,alt={Figure 5. Median loss(edit) minus loss(verbatim), in nats per token, by copy count of the original, for three OLMo-2 sizes. The first three groups are the mid-book bank. The fourth, past the dotted line, is the famous bank, a different set of sentences with far higher counts. Bars are 95 percent bootstrap intervals over pairs. Within noise, the gap does not depend on copy count between 1 and 999, and at 7B and 13B it is larger above that.}]{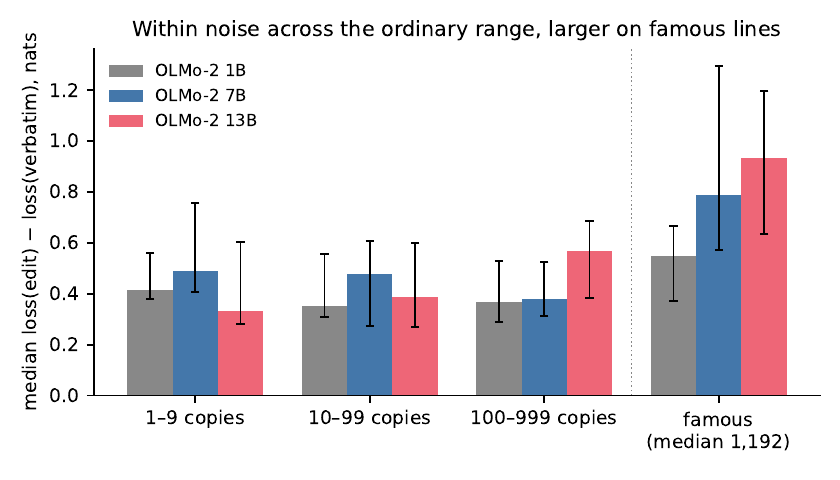}}
\caption{Median loss(edit) minus loss(verbatim), in nats per
token, by copy count of the original, for three OLMo-2 sizes. The first
three groups are the mid-book bank. The fourth, past the dotted line, is
the famous bank, a different set of sentences with far higher counts.
Bars are 95 percent bootstrap intervals over pairs. Within noise, the
gap does not depend on copy count between 1 and 999, and at 7B and 13B
it is larger above that.}
\end{figure}

\section{7. The control set does the
rest}\label{the-control-set-does-the-rest}

The famous bank carries two kinds of control for the same twelve
members. One is the same-source edit of Section 3.2. The other is prose
composed for this study, ordinary modern sentences of similar length,
which no corpus contains. The loss test then has two jobs on the same
twelve members. Asked to tell the members from their one-word edits, it
scores 0.83 (interval 0.65 to 0.98). Asked to tell the same members from
the composed sentences, it scores 0.94 (interval 0.85 to 1.00). The
members did not change. The 0.11 that separates the two numbers comes
from the controls. The composed sentences cost the model more than the
edits do, with a median loss of 3.57 nats per token against 2.69 for the
edits and 1.90 for the members. Two differences could produce that. One
is the register of plain modern prose against the members' older
literary English. The other is that each edit keeps all but one or two
words of a famous line, and this bank cannot separate the two. The
intervals overlap at this n, so treat this as a point estimate rather
than a test. Changing only the control set moved the estimate by about a
tenth of the AUC range, in a bank built to avoid exactly that. The same
two comparisons at 13B separate perfectly. Every famous member costs
less than every one-word edit and less than every composed sentence, so
both scores are 1.000 and the bootstrap interval collapses to a point.
Read that as a ceiling rather than as a stronger version of the 1B
result. This bank cannot rank the two control types at 13B, because both
already separate completely, so the 0.11 stays a 1B measurement.

Figure 6 puts the two effects in one picture. Count moves the number
from 0.60 to 0.83, and the control set moves it again to 0.94. That
second move is the distribution-shift complaint of Meeus et
al.~{[}2024{]} with a measured size attached.

\begin{figure}
\centering
\pandocbounded{\includegraphics[keepaspectratio,alt={Figure 6. AUC of the loss test on OLMo-2 1B for three member-and-control pairings. Left: mid-book members against same-source edits. Middle: famous members against same-source edits. Right: the same famous members against composed-prose controls. Bars are 95 percent bootstrap intervals.}]{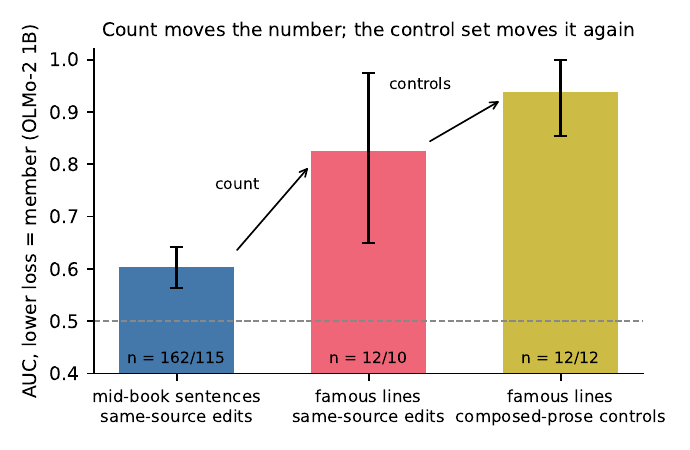}}
\caption{AUC of the loss test on OLMo-2 1B for three
member-and-control pairings. Left: mid-book members against same-source
edits. Middle: famous members against same-source edits. Right: the same
famous members against composed-prose controls. Bars are 95 percent
bootstrap intervals.}
\end{figure}

\section{8. The pincer}\label{the-pincer}

Put Sections 4 and 5 side by side.

Loss-based evidence of membership becomes strong only when a sentence
has been duplicated on the order of a thousand times or more. Among
literary sentences, the ones duplicated that often are the famous ones:
opening lines, quotations, verses. And famous sentences are famous
everywhere. Across the twelve famous lines of the positive-control bank,
the ratio of OLMo-mix copies to Pile copies varies only six-fold between
the tenth and ninetieth percentiles, against twenty-seven-fold for
mid-book sentences, and every famous line sits on the same side,
commoner in OLMo-mix. In log10 the famous span is 0.78, from 0.79 to
1.57, against 1.44 for mid-book. So at the counts where exposure is
detectable, our two corpora do not let us vary exposure while holding
the sentence fixed, and the detector's signal cannot be separated from
whatever else makes a sentence famous.

Below those counts the corpora do disagree, so exposure can be varied,
but there the trace is faint. It is about $-$0.08 in rank correlation at
7B and weaker at 13B, under one percent of the variance.

Those two findings are the pincer. Membership evidence from loss is
either weak or confounded, and the transition between the two regimes is
not a place where a careful test can stand. For the models and the text
we tested, a detector reporting strong membership at ordinary
duplication levels is reporting something other than membership, and
Sections 6 and 7 name the two most likely candidates.

Three experiments would break the pincer. A model large enough to carry
a strong sentence-level trace at tens of copies would open the low
regime, and since memorization scales with size, 13B is not the end of
the ladder. A pair of documented corpora that disagree about famous text
would open the high regime; we know of no such pair among public
releases, but one could be built. And a controlled injection of unique
sequences during pretraining, the design recommended by Meeus et
al.~{[}2024{]}, would sidestep the natural experiment entirely; it costs
a pretraining run, which is why the natural experiment was worth doing
first.

The first of those three has already begun to move, in our own data, and
against us. Every model reads the same twelve famous sentences, so their
fame is held exactly fixed across the size ladder while the capacity to
have memorized them is not. Pool each model's paired gaps, mid-book and
famous together, and ask whether the gap tracks copy count. At 13B it
does, at a rank correlation of 0.206 with a permutation p of 0.039. At
the two smaller models it does not, at $-$0.042 and 0.036. That reading
never passes through a corpus comparison, and because every model reads
the same twelve famous lines, fame alone cannot explain why only 13B
shows it. One model, one bank of twelve, and the 13B scores are 8-bit,
so it settles nothing on its own. It is the first crack in our own
pincer, and it is where we would push.

\section{9. Related work}\label{related-work}

Membership inference on language models is usually framed as an attack
and evaluated as a classifier {[}Shi et al., 2024; Mattern et al.,
2023{]}. Duan et al.~{[}2024{]} evaluated the major attacks across the
Pythia suite and found them near chance, using bloom-filter membership
over the Pile with no per-item counts; their Table 2 shows that changing
how non-members are constructed changes the result. Meeus et
al.~{[}2024{]} reviewed the evaluation practice and identified post-hoc
set construction as the central weakness; Das et al.~{[}2024{]} showed
that blind baselines match published attacks on exactly the shifts that
practice introduces; Zhang et al.~{[}2024{]} argued from the same
premises that a membership test cannot prove training on a document.
This paper is in that line, and adds two things it lacked. The first is
the copy count as a measured dose, and the second is a design that
varies the dose within a fixed sentence.

Carlini et al.~{[}2022{]} showed that verbatim memorization, measured by
extraction, scales with model size, duplication, and prompt length,
using duplicate counts over the Pile. Our copy counts are the same
quantity read from a different index. What differs is the measurement on
the model side. Extraction asks whether the model can reproduce a
sequence, while we ask whether its loss on the sequence carries any
trace at all, which is the weaker and more relevant question for
membership. Kandpal et al.~{[}2022{]} found that detectors of memorized
sequences are near chance on sequences that appear only once, and
Section 4 measures where loss starts to do better.

The two-family design has a distant relative in reference-model attacks,
which compare a target model's loss to that of a model trained on
similar data {[}Mireshghallah et al., 2022; Carlini et al., 2021{]}.
Those attacks use the reference as a fluency baseline. We use it as a
second exposure condition with its own measured dose, which is only
possible because both corpora are public.

\section{10. Limitations, and what the release
enables}\label{limitations-and-what-the-release-enables}

The measurements are limited in five stated ways. Model size stops at
13B, and the 13B-class runs used 8-bit precision. The text is English
literary prose from six books, in sentences of 10 to 16 words, plus
twelve famous lines. Copy counts are exact-match floors, and on the OLMo
side they omit an annealing stage. There is no injected-sequence
condition. And the famous-line positive control is twelve sentences, so
every interval that rests on it is correspondingly wide. A larger famous
bank carrying both counts would narrow them and is inexpensive to build.

Each of these is a reason the numbers could move. At the sizes tested,
none of them changes the shape of the two jaws, which are set by
properties of public corpora and of exact matching rather than by our
choices. Precision is the exception worth naming, because quantization
noise at 13B could flatten a size trend, and Section 5.2 reads the flat
13B trend with caution for that reason. Beyond the sizes tested, the
first limitation is also the first way to break the pincer named in
Section 8, since a larger model is the experiment these limits point at.
A larger bank would narrow the intervals in Figures 2 and 6 and, with
more books, decide whether the between-book pattern in Figure 3 is taste
or memory.

\begin{sloppypar}
We release the sentence banks with both copy counts, the scoring and
analysis code, and the figure scripts, under an MIT licence. For an even
number of values, the edit-gap medians, the copy-count medians that
place Figure 2's points, the book medians behind the between-book
correlations, and the famous bank's median copy count take the upper of
the two middle values, as the released script does. The repository is
\mbox{\url{https://github.com/IamArmanNikkhah/detectable-only-where-confounded}}.
Labelling a new sentence costs one query to a free index, and scoring a
bank under a 7B model costs seconds of GPU time. The design generalizes
to any pair of models whose corpora are indexed, and infini-gram already
indexes several.
\end{sloppypar}

\section{References}\label{references}

Biderman, S., et al.~(2023). Pythia: A suite for analyzing large
language models across training and scaling. arXiv:2304.01373.

Carlini, N., et al.~(2021). Extracting training data from large language
models. USENIX Security. arXiv:2012.07805.

Carlini, N., Ippolito, D., Jagielski, M., Lee, K., Tramèr, F., and
Zhang, C. (2022). Quantifying memorization across neural language
models. arXiv:2202.07646.

Das, D., Zhang, J., and Tramèr, F. (2024). Blind baselines beat
membership inference attacks for foundation models. arXiv:2406.16201.

Duan, M., Suri, A., Mireshghallah, N., Min, S., Shi, W., Zettlemoyer,
L., Tsvetkov, Y., Choi, Y., Evans, D., and Hajishirzi, H. (2024). Do
membership inference attacks work on large language models? COLM.
arXiv:2402.07841.

Gao, L., et al.~(2020). The Pile: An 800GB dataset of diverse text for
language modeling. arXiv:2101.00027.

Kandpal, N., Wallace, E., and Raffel, C. (2022). Deduplicating training
data mitigates privacy risks in language models. ICML. arXiv:2202.06539.

Liu, J., Min, S., Zettlemoyer, L., Choi, Y., and Hajishirzi, H. (2024).
Infini-gram: Scaling unbounded n-gram language models to a trillion
tokens. arXiv:2401.17377.

Mattern, J., Mireshghallah, F., Jin, Z., Schölkopf, B., Sachan, M., and
Berg-Kirkpatrick, T. (2023). Membership inference attacks against
language models via neighbourhood comparison. arXiv:2305.18462.

Meeus, M., Shilov, I., Jain, S., Faysse, M., Rei, M., and de Montjoye,
Y.-A. (2024). SoK: Membership inference attacks on LLMs are rushing
nowhere (and how to fix it). arXiv:2406.17975.

Mireshghallah, F., et al.~(2022). Quantifying privacy risks of masked
language models using membership inference attacks. EMNLP.
arXiv:2203.03929.

OLMo Team (2025). 2 OLMo 2 Furious. arXiv:2501.00656.

Shi, W., et al.~(2024). Detecting pretraining data from large language
models. ICLR. arXiv:2310.16789.

Zhang, J., Das, D., Kamath, G., and Tramèr, F. (2024). Membership
inference attacks cannot prove that a model was trained on your data.
arXiv:2409.19798.

\section{Appendix A. A method that did not
work}\label{appendix-a.-a-method-that-did-not-work}

The project began as a test of a different membership signal, namely how
hard a sentence is to unlearn by gradient ascent, on the hypothesis that
memorized text resists forgetting. Under plain SGD ascent on OLMo-2 1B,
with a fixed step budget chosen below the point at which unrelated
control text begins to degrade, the resistance measure carried no
information beyond the starting loss. Its AUC after regressing out
baseline loss was 0.54 on the famous bank, read at a 44-step budget, the
largest at which no control text had begun to degrade, and 0.40 on a
100-sentence subset of pass 3 at 46 steps. We do not report the method
further. The banks and the two-corpus design were built to give it a
fair test, and they turned out to be the result.

\section{Appendix B. Pass-3 scores against the edits
alone}\label{appendix-b.-pass-3-scores-against-the-edits-alone}

Section 3.3 notes that 13 of the 115 pass-3 non-members are sampled book
sentences that the index returns zero times, not one-word edits. At the
median they cost OLMo-2 1B less than the members do, 3.64 nats per token
against 4.23 for the members and 4.50 for the edits, which is why
leaving them out raises every score. Against the 102 edits alone, the
loss test scores 0.63 on the full bank (95 percent bootstrap interval
0.59 to 0.67), where Section 4 reports 0.60 with all 115. By copy band
the scores are 0.69, 0.62 and 0.60 for members with 1 to 9, 10 to 99 and
100 to 999 copies, against 0.66, 0.59 and 0.57 with all 115. Every score
rises by about 0.03. The pattern of Section 4 holds, because the bands
still overlap and the lowest band still scores highest.

\end{document}